\documentclass[10pt,twocolumn,letterpaper]{article}

\usepackage[pagenumbers]{cvpr} %

\newcommand{\name}{Sat2Sound}
\usepackage{fontawesome5,array}
\usepackage[accsupp]{axessibility}  %

\makeatletter
\def\blfootnote{\xdef\@thefnmark{}\@footnotetext}
\makeatother

\definecolor{cvprblue}{rgb}{0.21,0.49,0.74}
\usepackage[pagebackref,breaklinks,colorlinks,allcolors=cvprblue]{hyperref}
\usepackage{multirow}
\usepackage{pifont}
\def\paperID{56} %
\def\confName{EarthVision}
\def\confYear{2026}

\title{Sat2Sound: A Unified Framework for Zero-Shot
Soundscape Mapping}

\author{
Subash Khanal$^{1}$\textsuperscript{\faEnvelope}, Srikumar Sastry$^1$, Aayush Dhakal$^1$, Adeel Ahmad$^{1,2}$\\
Abby Stylianou$^3$, Nathan Jacobs$^1$\\
$^1$Washington University in St.\ Louis \quad
$^2$Taylor Geospatial \quad
$^3$Saint Louis University\\
}

\begin{document}
\maketitle
\blfootnote{\textsuperscript{\faEnvelope}Corresponding Author: \texttt{k.subash@wustl.edu}}
\begin{abstract}
We present \name, a unified multimodal framework for geospatial soundscape understanding, designed to predict and map the distribution of sounds across the Earth's surface. Existing methods for this task rely on paired satellite images and geotagged audio samples, which often fail to capture the full diversity of sound at a location. Sat2Sound overcomes this limitation by augmenting datasets with semantically rich, vision-language model-generated soundscape descriptions, which broaden the range of possible ambient sounds represented at each location. Our framework jointly learns from audio, text descriptions of audio, satellite images, and synthetic image captions through contrastive and codebook-aligned learning, discovering a set of ``soundscape concepts'' shared across modalities, enabling hyper-localized, explainable soundscape mapping. \name~achieves state-of-the-art performance in cross-modal retrieval between satellite image and audio on the GeoSound and SoundingEarth benchmarks. Finally, by retrieving detailed soundscape captions that can be rendered through text-to-audio models, \name~enables location-conditioned soundscape synthesis for immersive and educational applications, even with limited computational resources. Our code and models are available at \url{https://github.com/mvrl/sat2sound}.

\end{abstract}
    
\section{Introduction}
\label{intro}

Imagine exploring our planet and listening to the sounds of a specific place, or creating a map that highlights locations that resemble your imagined soundscape. This is the essence of soundscape mapping—predicting and visualizing the ambient acoustic environment of any location on Earth. Such capability supports immersive geospatial audio exploration, biodiversity and urban noise monitoring, sound-conscious urban design~\cite{kotian2024measuring, aletta2019associations}, and even consumer applications, helping real estate buyers and tourists find locations that match their acoustic preferences~\cite{morano2021economic}.

Recognizing the value of these capabilities, recent efforts have focused on developing frameworks for soundscape mapping~\cite{khanal2023soundscape,khanal2024psm}. These frameworks represent locations using satellite images and learn a trimodal embedding space that links the satellite image, audio, and textual descriptions of the audio. Datasets used to train these frameworks contain geotagged audio recordings collected from various crowdsourced platforms. However, these audio samples typically fail to capture the full diversity of sound sources at the recorded locations. We address this limitation by introducing a data-driven augmentation strategy: querying a powerful vision-language model (VLM) to describe the soundscape implied by each satellite image. These automatically generated captions enumerate plausible ambient sources, providing semantic coverage beyond what any single audio sample can capture.

Building on these enriched soundscape descriptions, we propose \name, a unified multimodal representation-learning framework for geospatial soundscape understanding. \name~jointly aligns audio, textual descriptions, satellite imagery, and synthetic VLM captions through contrastive learning over a shared codebook of soundscape concepts. Each codebook entry acts as a shared anchor across modalities, and during training, the model learns to associate subsets of entries with recurring acoustic and visual patterns (e.g., urban traffic or bird chorus). By learning these shared soundscape concepts, \name~shifts from global representations toward a structured, concept-level alignment between image regions and characteristic acoustic patterns. This intuition draws on ideas from FILIP~\cite{yao2021filip} and FDT~\cite{chen2023revisiting}, where token-level and codebook-based alignment improved interpretability in image–text models; here, we generalize that insight to a multimodal, geospatial setting connecting satellite imagery, audio, and text.

This approach contrasts with prior soundscape mapping frameworks such as GeoCLAP~\cite{khanal2023soundscape} and PSM~\cite{khanal2024psm}, which model each location with a single global embedding, limiting their ability to represent the compositional and overlapping nature of real-world soundscapes. \name~instead learns a representation that captures these mixtures through local codebook-based alignment, adding in metadata conditioning (location, time, and source), and VLM-augmented captions that provide contextual grounding and semantic diversity. Together, these elements result in a framework that connects satellite imagery, ambient audio, and textual descriptions within a shared, interpretable space, supporting both localized soundscape mapping and cross-modal retrieval among all three modalities. 

Although trained discriminatively, \name's retrieval capabilities naturally enable generative applications. Retrieved soundscape captions can be rendered into audio using state-of-the-art text-to-audio models, producing realistic, location-conditioned sounds. These captions, generated by a vision–language model, describe the full range of plausible ambient sources at each site—far richer than the limited metadata or short annotations in existing soundscape datasets. However, directly generating audio everywhere remains computationally expensive and impractical for low-resource deployments such as mobile monitoring, educational outreach, or interactive exhibits. Instead, \name\ supports a retrieval-based strategy: precompute synthetic audio for all captions once, and at inference time, retrieve the most representative sound for a given image. This eliminates heavy generative inference, relying only on efficient retrieval over high-quality embeddings, and enables interactive, globally deployable soundscape synthesis with minimal compute.

In summary, \name~introduces:
\begin{itemize}
   \item a unified multimodal framework for geospatial soundscape mapping that integrates satellite imagery, audio, and both human- and VLM-generated captions that capture a rich and compositional description of ambient acoustic environments, achieving state-of-the-art results on multiple benchmarks;
   \item learnable codebook that represents a finite set of soundscape concepts shared across modalities, enhancing alignment between image patches and soundscape concepts;
   \item a retrieval-based synthesis approach that enables low-cost, location-aware soundscape generation, even in resource-limited environments.
\end{itemize}

\section{Related Work}
\textbf{Audio Visual Learning:} There is a strong semantic relationship between the acoustic and visual signals in a given audio-visual sample. Several studies~\cite{chen2023revisiting, khanal2023soundscape, khanal2024psm, salem2018multimodal, zeng2023learning, sheffer2023hear, Sung-Bin_2023_CVPR, gao2024audio} have leveraged this relationship to develop powerful audio-visual models. In the context of conditional audio generation, recent works~\cite{sheffer2023hear, wang2024v2a} have utilized existing foundational models to generate semantically meaningful audio from input images, while~\cite{zhang2025audio, Sung-Bin_2023_CVPR} proposed models that generate images from audio.
For soundscape mapping, recent works~\cite{khanal2023soundscape, khanal2024psm} explored this domain by learning global embeddings between overhead images and geotagged sounds, enabling zero-shot retrieval across locations. However, these global representations are limited in their ability to model compositional and overlapping sound sources or to explain why an image corresponds to a given sound. \name~extends this line of work by incorporating local codebook-based alignment that grounds interpretable soundscape concepts across modalities and by incorporating VLM-generated captions to allow for richer notions of the sounds that are likely to exist for a given satellite image and human-generated sound annotation.
\newline
\textbf{Contrastive Learning:} Contrastive learning is an effective strategy to learn a shared embedding space between multiple modalities~\cite{radford2021learning, li2021align, yucoca, vivanco2024geoclip, khanal2024psm, girdhar2023imagebind}. The shared embedding space can be either deterministic or probabilistic. For example, \cite{radford2021learning} used contrastive learning to align large-scale image-text pairs, learning a deterministic image-text embedding space.
Some recent works have proposed learning a probabilistic embedding space~\cite{chun2021probabilistic, chunimproved} between modalities. While most of these methods focus on learning a single representation per sample to encourage global alignment between modalities, some recent works, such as FILIP~\cite{yao2021filip} and FDT~\cite{chen2023revisiting} have adapted contrastive learning to encourage local alignment between image and text. MGA-CLAP~\cite{li2024advancing} takes a similar approach to FDT and learns alignment between audio and text. Motivated by these approaches, we also train our framework by adapting a contrastive learning method~\cite{chen2023revisiting}, which fosters local alignment between satellite images and their soundscape descriptions.
\newline
\textbf{Discrete Representation Learning:} Discrete representation learning focuses on learning a discrete latent space (codebook) composed of a fixed number of concepts. These discrete latent concepts are typically learned as intermediate weights within encoder-decoder frameworks, such as the Vector-Quantized Variational Autoencoder (VQ-VAE)~\cite{van2017neural}. VQ-VAE-style codebook learning has been applied to various conditional generation tasks , including text-to-image generation~\cite{gu2022vector}, image-to-audio generation~\cite{iashin2021taming}, and text-to-audio generation~\cite{yang2023diffsound}. 
Beyond generative models, codebook learning has also been adopted in different cross-modal representation learning frameworks~\cite{liu2022cross,xia2024achieving,li2024advancing,chen2023revisiting}. Drawing inspiration from these works, we design our framework to learn a discrete set of soundscape concepts shared across our modalities: images, text, and audio.

\section{Method}
\begin{figure*}[ht]
    \centering
    \includegraphics[width=\textwidth]{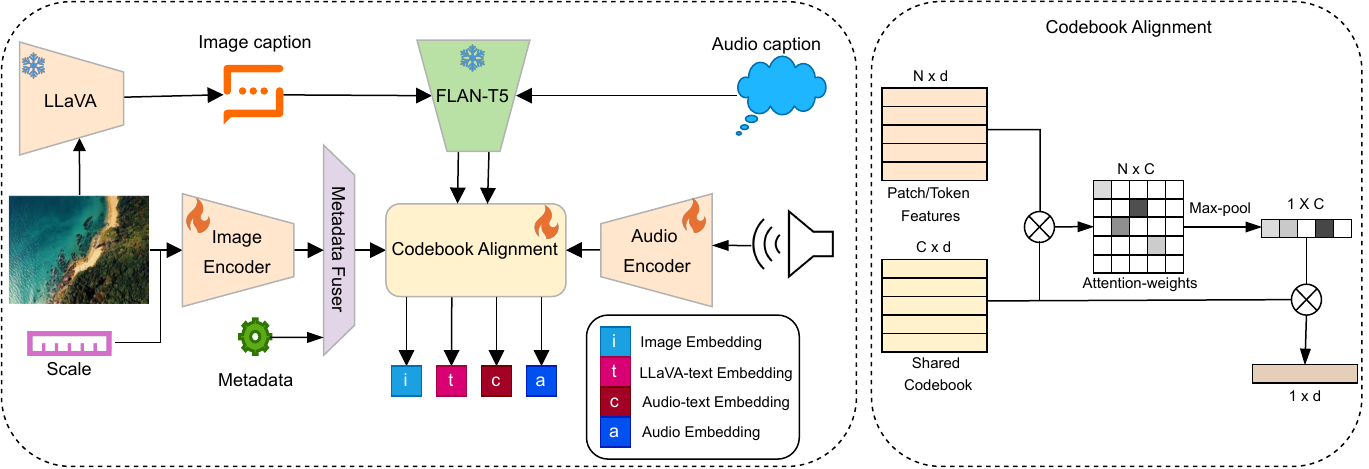}
    \caption{Sat2Sound framework learns a shared multimodal embedding space between satellite images, audio, audio captions, and image captions. Modality-specific encoders generate token embeddings for each modality, which are aligned into a shared codebook through an attention-score-based concept aggregation process.}
    \label{fig:framework}
\end{figure*}
This section describes our proposed framework: \name, a multimodal representation learning framework for soundscape mapping. 

Figure \ref{fig:framework} provides an overview of \name, which incorporates encoders for satellite images, audio, and text. \name\ is trained on two types of text: audio captions that describe the semantics of specific geotagged audio clips, as well as synthetically generated textual descriptions that describe the potential soundscape for a given satellite image. We additionally incorporate associated metadata for each sample (audio source, audio caption source, location, month, and time) and leverage the multiscale nature of satellite imagery to enable metadata-aware, multi-scale soundscape mapping within our framework. %

{
\begin{figure}[ht]

  \centering
  {
  \tiny
  \begin{tabular}{@{}p{1in}@{\hspace{6pt}}p{\dimexpr\columnwidth-1in-6pt\relax}@{}}
    \begin{minipage}[t]{1in}\vspace{0pt}
      \includegraphics[width=1in]{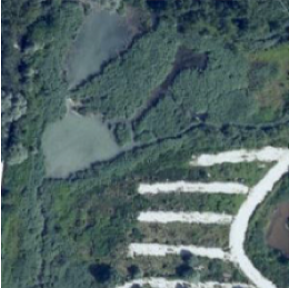}
    \end{minipage}
    &
    \vspace{2em}\begin{minipage}[t]{\dimexpr\columnwidth-1in-6pt\relax}\vspace{0pt}\raggedright\scriptsize
      \textbf{Original Audio Caption:} ``birds are chirping''\par
      \vspace{1em}\textbf{Synthetic Caption:} ``From the location captured in the aerial view image, we can expect to hear the sounds of birds chirping, leaves rustling, and the gentle flow of water from the pond.''
    \end{minipage}
    \\[8pt]
    \begin{minipage}[t]{1in}\vspace{0pt}
      \includegraphics[width=1in]{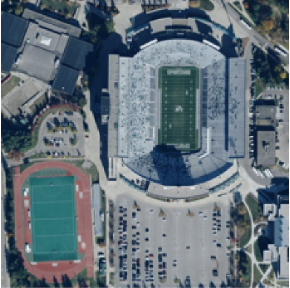}
    \end{minipage}
    &
    \begin{minipage}[t]{\dimexpr\columnwidth-1in-6pt\relax}\vspace{0pt}\raggedright\scriptsize
      \textbf{Original Audio Caption:} ``A victory chant is being made.''\par
      \vspace{1em}\textbf{Synthetic Caption:} ``From the location captured in the aerial view image, we can expect to hear the sounds of cars driving in the parking lot, people walking around, and possibly the noise from the stadium or arena, depending on the time of day and the event taking place.''
    \end{minipage}
    \\[8pt]
    \begin{minipage}[t]{1in}\vspace{0pt}
      \includegraphics[width=1in]{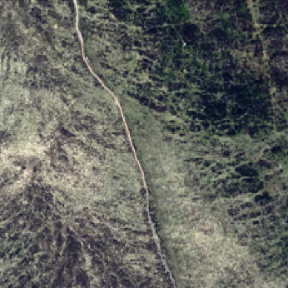}
    \end{minipage}
    &
    \vspace{2em}\begin{minipage}[t]{\dimexpr\columnwidth-1in-6pt\relax}\vspace{0pt}\raggedright\scriptsize
      \textbf{Original Audio Caption:} ``Light breeze in a field.''\par
      \vspace{1em}\textbf{Synthetic Caption:} ``From the location captured in the aerial view image, we can expect to hear the sounds of wind rustling through the tall grass, birds chirping, and possibly the occasional rustling of leaves or branches.''
    \end{minipage}
  \end{tabular}
  }
  \caption{Comparison of original audio captions and synthetic LLaVA captions generated from satellite imagery.}
  \label{fig:synthetic_data}
\end{figure}
}

\subsection{Synthetic Textual Descriptions}\label{sec:synthetic_text}

Our training data consists of samples from GeoSound~\cite{khanal2024psm} and SoundingEarth~\cite{heidler2023self}. These datasets include geotagged audio samples and corresponding audio captions. These captions can either come from the user-uploaded textual descriptions or be generated using recent SOTA audio-to-text generation models such as Pengi~\cite{deshmukh2023pengi} or Qwen-Audio~\cite{Qwen-Audio} (the final caption is selected based on its CLAP score~\cite{wu2023large} with the audio). While these captions provide useful context for the sound at a given location, they often capture only a narrow instance of what might be heard at that location based on the single audio sample being annotated. For example, a satellite image of an urban intersection might have a label of ``honking car horns'' but also correspond to sounds of pedestrians, buses, and construction.

To expand the range of acoustic contexts represented in training, we generate synthetic textual descriptions for each satellite image using the vision–language model LLaVA \cite{liu2024visual}. We prompt the model with ``What types of sounds can we expect to hear from the location captured by this aerial view image? Describe in up to two sentences.'' and obtain compositional, multi-source descriptions that reflect plausible ambient soundscapes rather than isolated events. Figure~\ref{fig:synthetic_data} illustrates this contrast: each column shows a satellite image, its corresponding annotation from GeoSound or SoundingEarth, describing a single recorded sound, and the richer LLaVA-generated caption enumerating multiple likely sources at that location.

The synthetic soundscape captions serve as additional image-level text supervision, used alongside the original audio captions during multimodal training. By combining the two, \name\ learns to align visual and acoustic information not only for the specific sounds present in the dataset but also for the broader distribution of sounds that could co-occur at similar locations.

\subsection{Encoding Modalities}
Each sample ($X$) used in our training consists of geotagged audio $X^{a}$, its corresponding audio caption $X^{c}$, a satellite image at a scale $s$, taken at the audio-recording location, $X_{s}^{i}$, and the associated image caption $X^{t}$. Modality-specific encoders, $E_{audio}$, $E_{text}$, and $E_{image}$, are used to obtain patch/token-level representations, each projected into a $d$-dimensional embedding space: $h^{a} \in \mathbb{R}^{N^a \times d}$, $h^{c} \in \mathbb{R}^{N^c \times d}$, $h^{i,s} \in \mathbb{R}^{N^i \times d}$, and $h^{t,s} \in \mathbb{R}^{N^{t,s} \times d}$ for audio, audio caption, image at scale $s$, and image caption, respectively. Here, $N^a$ represents the number of frames in the audio feature, $N^c$ is the number of tokens in the audio caption, $N^i$ is the number of patches in the satellite image, and $N^{t,s}$ is the number of tokens in the image caption:
{
\begin{equation}\label{eq:1}
h^{y} = E_{y}(X^{y}),
\end{equation}
}
where $y \in \{\text{audio}, \text{ audio caption}, \text{image}, \text{image caption}\}$ and $E_{y}$ is the modality-specific encoder.

To learn a metadata-aware embedding space, we adopt an early-fusion strategy, where we combine $5$ metadata components (geolocation, month, hour, audio source, and audio caption source) with the patch embeddings for the satellite image obtained from Equation~\ref{eq:1}. Specifically, each metadata component is first embedded into $d$-dimensional representations using shallow linear layers, and these representations are concatenated with the image patch embeddings, along the patch dimension. The concatenated input is then passed through a transformer module to obtain a metadata-conditioned satellite image representation:
{
\begin{equation}\label{eq:2} 
h^{i'} = E_{meta}(h^{i}, \text{metadata})
\end{equation}
}
where $E_{meta}$ represents the metadata fusion module, and $h^{i'} \in \mathbb{R}^{(N^i + 5) \times d}$ is the resulting metadata-conditioned satellite image patch embeddings.
\subsection{Codebook Alignment}

Once the modality-specific encoders compute the patch/token embeddings for each modality, the next step is to project them into a shared embedding space. To achieve this, we adopt a discrete representation learning strategy~\cite{chen2023revisiting, li2024advancing}, which learns a shared codebook, $C \in \mathbb{R}^{M \times d}$, representing $M$ soundscape concepts shared across the modalities. 

To illustrate the process, let us consider the case where the current modality of interest is the satellite image. Let $p^{i}_{j} \in \mathbb{R}^{d}$ denote the $j$-th patch embedding for sample $i$, and let $C_m \in \mathbb{R}^{d}$ be the $m$-th codebook token. For each codebook token, we compute a relevance score by taking the inner product with every patch embedding and selecting the maximum value across all patches $(j = 1,\ldots,N^{i})$:
{
\begin{equation}
r_m^i = \max_{j} \left( p^{i}_{j} \cdot C_m \right),
\end{equation}
}
The resulting relevance scores are then normalized using a Softmax function to obtain attention weights:
{
\begin{equation}
\label{eq:4}
w_m^{i} = \frac{\exp(r_m^{i})}{\sum_{n=1}^{M} \exp(r_n^{i})}.
\end{equation}
}
Following \cite{chen2023revisiting}, we further apply a Sparsemax function \cite{martins2016softmax} to these weights to obtain a sparser distribution, which reduces noise and improves interpretability for grounding.

Using the same process, normalized attention weights for other modalities—audio caption ($w_m^c$), image caption ($w_m^t$), and audio ($w_m^a$)—are computed using the token/frame embeddings from their respective encoders and the shared codebook $C$.
These attention weights enable each modality to dynamically attend to relevant codebook tokens, facilitating cross-modal alignment in the shared embedding space. Although the learning objective does not explicitly enforce distinct semantics for each codebook entry, the combination of contrastive alignment and Sparsemax attention encourages specialization, leading to interpretable codebook of soundscape concepts in practice (see Section~\ref{sec:codebook_alignment} and Figure \ref{fig:codebook_groups_examples} for qualitative examples).

Finally, the pooled embeddings for each modality ($y$) are obtained as a weighted sum of all the codebook concepts:
{
\begin{equation}
\label{eq:5}
   f^y = \sum_{m=1}^M w_m^y \cdot C_m, \quad y \in \{i,a,c,t\},
\end{equation}
}
where $f^i$, $f^a$, $f^c$, and $f^t$ are the codebook-aligned embeddings for the image ($i$), audio ($a$), audio caption ($c$), and image caption ($t$) respectively.

\subsection{Multimodal Contrastive Learning}
Finally, the codebook-aligned embeddings obtained from Equation~\ref{eq:5} are used in our multimodal contrastive learning framework. For a pair of modalities ($u$,$v$), we use the InfoNCE loss~\cite{infonce,radford2021learning} which is defined as follows:
{
\begin{equation}
\label{eq:6}
\begin{aligned}
\mathcal{L}_{u,v} = 
& -\frac{1}{2B} \Bigg( 
\sum_{n=1}^B \log 
\frac{\exp\left((f_{n}^u \cdot f_{n}^v)/\tau_{uv}\right)}
     {\sum_{s=1}^B \exp\left((f_{s}^u \cdot f_{s}^v)/\tau_{uv}\right)} 
\\
& \quad +
\sum_{n=1}^B \log 
\frac{\exp\left((f_{n}^v \cdot f_{n}^u)/\tau_{uv}\right)}
     {\sum_{s=1}^B \exp\left((f_{s}^v \cdot f_{s}^u)/\tau_{uv}\right)} 
\Bigg)
\end{aligned}
\end{equation}
}
where $\tau_{uv}$ is the learnable temperature parameter and $B$ is the batch size during training.

\begin{table*}[!ht]
\centering
\small
\setlength{\tabcolsep}{6pt}
\caption{Cross-modal retrieval performance comparison of \name~across different datasets.}
\label{tab:main}
\begin{tabular}{c|l|c|cc|cc}
\hline
\multirow{2}{*}{Dataset} & \multirow{2}{*}{Method} & \multirow{2}{*}{Metadata} &
\multicolumn{2}{c|}{Image-to-Audio} & \multicolumn{2}{c}{Audio-to-Image} \\
\cline{4-7}
 &  &  & R@10\% & MR & R@10\% & MR \\
\hline
\multirow{6}{*}{GeoSound-Bing}
  & GeoCLAP \cite{khanal2023soundscape} & \multirow{3}{*}{\ding{55}}
      & 0.399 & 1500 & 0.403 & 1464 \\
  & PSM \cite{khanal2024psm} &  & 0.423 & 1401 & 0.428 & 1344 \\
  & Ours &  & \textbf{0.534} & \textbf{872} & \textbf{0.535} & \textbf{850} \\
\cline{2-2}\cline{3-7}
  & PSM \cite{khanal2024psm} & \multirow{2}{*}{\checkmark} 
      & 0.828 & 261 & 0.829 & 248 \\
  & Ours &  & \textbf{0.871} & \textbf{168} & \textbf{0.875} & \textbf{164} \\
\hline
\multirow{6}{*}{GeoSound-Sentinel}
  & GeoCLAP \cite{khanal2023soundscape} & \multirow{3}{*}{\ding{55}}
      & 0.459 & 1179 & 0.465 & 1141 \\
  & PSM \cite{khanal2024psm} &  & 0.474 & 1101 & 0.485 & 1061 \\
  & Ours &  & \textbf{0.549} & \textbf{802} & \textbf{0.556} & \textbf{778} \\
\cline{2-2}\cline{3-7}
  & PSM \cite{khanal2024psm} & \multirow{2}{*}{\checkmark}
      & 0.802 & 294 & 0.804 & 283 \\
  & Ours &  & \textbf{0.868} & \textbf{191} & \textbf{0.872} & \textbf{183} \\
\hline
\multirow{6}{*}{SoundingEarth}
  & GeoCLAP \cite{khanal2023soundscape} & \multirow{3}{*}{\ding{55}}
      & 0.454 & 667 & 0.449 & 694 \\
  & PSM \cite{khanal2024psm} &  & 0.514 & 547 & 0.518 & 543 \\
  & Ours &  & \textbf{0.570} & \textbf{438} & \textbf{0.562} & \textbf{463} \\
\cline{2-2}\cline{3-7}
  & PSM \cite{khanal2024psm} & \multirow{2}{*}{\checkmark}
      & 0.563 & 454 & 0.569 & 447 \\
  & Ours &  & \textbf{0.626} & \textbf{358} & \textbf{0.621} & \textbf{372} \\
\hline
\end{tabular}\vspace{-1em}
\label{tab:consolidated_results}
\end{table*}

Given the one-to-many nature of satellite–soundscape correspondence, batches inherently contain pseudo-positives—samples labeled as negatives but semantically similar to the true match. Following~\cite{chunimproved,khanal2024psm}, we incorporate these samples into the contrastive loss. A pseudo-positive is defined as any within-batch sample whose similarity to the query is greater than or equal to the ground-truth similarity. We adopt this soft-matching strategy and modify our overall objective as follows:
{
\begin{equation}
\label{eq:7}
\mathcal{L}_{u,v}^{\dagger} = \mathcal{L}_{u,v} + \alpha \cdot \mathcal{L}_{u,v}^{\text{pseudo}},
\end{equation}
}
where $\mathcal{L}_{u,v}^{\text{pseudo}}$ is the contrastive loss computed by treating pseudo-positive samples as additional positives within the batch, and $\alpha$ controls the strength of this term. Using Equation~\ref{eq:7}, we compute the total loss for the following four modality pairs
: image and audio ($\mathcal{L}_{i,a}^{\dagger}$), image and audio caption ($\mathcal{L}_{i,c}^{\dagger}$), audio and audio caption ($\mathcal{L}_{a,c}^{\dagger}$), and image and image caption ($\mathcal{L}_{i,t}^{\dagger}$).  Finally, our trimodal contrastive learning objective is formulated as:
{
\begin{equation}
\label{eq:8}
\mathcal{L}_{\text{tri}} = (\mathcal{L}_{i,a}^{\dagger} + \mathcal{L}_{i,c}^{\dagger} + \mathcal{L}_{a,c}^{\dagger})/3
\end{equation}
}
Additionally, we acknowledge that the audio-to-image retrieval task can be modified to a composed audio-to-image retrieval task. In this setting, an audio query is also accompanied by its caption. 
To explicitly embrace this scenario during training, we create a composed audio embedding by combining the audio embedding with the audio-caption embedding: $f^{a+c} = f^a + f^c$. The contrastive loss is then computed between this composed audio embedding and the image embedding ($\mathcal{L}_{i,a+c}^{\dagger}$). Finally, our overall objective function is as follows:
{
\begin{equation}
\label{eq:9}
\mathcal{L}_{total} = \mathcal{L}_{tri} + \mathcal{L}_{i,a+c}^{\dagger} + \mathcal{L}_{i,t}^{\dagger}
\end{equation}
}

\section{Evaluation \& Results}

We evaluate \name\ on both retrieval and mapping tasks. Quantitative retrieval metrics assess alignment quality across modalities, while qualitative soundscape maps illustrate how this alignment captures the geographic and semantic structure of environmental sounds. Together, these results demonstrate that \name~not only achieves strong cross-modal correspondence but also supports interpretable, large-scale, hyper-local mapping of soundscapes. Experimental details, including information on the datasets, input processing, encoders for each modality, and training hyperparameters, are provided in the supplemental materials.
\begin{table*}[t]
\centering
\small
\setlength{\tabcolsep}{4pt}
\caption{Image-Text retrieval results for different frameworks on GeoSound dataset with \textit{Bing} imagery.}
\label{tab:text_retrieval}
\begin{tabular}{l|c|c|cccc|ccc}
\hline
Method & Metadata & scale & I2T-R10\% & I2T-MR & T2I-R10\% & T2I-MR & METEOR & BLEU & BERT-F1 \\
\hline
Sat2Text & &1 & 0.904 & 164 & 0.912 & 138 & 0.697&0.524&0.926 \\
Sat2Sound& & 1 & 0.881 & 183 & 0.900 & 166 & 0.682&0.497&0.921\\
Sat2Sound&\checkmark& 1 & 0.908 & 160 & 0.914 & 136 & 0.688&0.520&0.925 \\
\hline
Sat2Text && 3 & 0.929 & 117 & 0.927 & 110 & 0.695&0.519&0.922 \\
Sat2Sound& & 3 & 0.918 & 134 & 0.921 & 120 & 0.675&0.491&0.917 \\
Sat2Sound& \checkmark& 3 & 0.940 & 106 & 0.938 & 97 & 0.689&0.513&0.920 \\
\hline
Sat2Text && 5 & 0.910 & 132 & 0.915 & 120 &0.651&0.480&0.916 \\
Sat2Sound& & 5 & 0.895 & 153 & 0.903 & 138&0.635&0.457&0.911 \\
Sat2Sound&\checkmark & 5 & 0.920 & 127 & 0.926 & 114 &0.650&0.468&0.914 \\
\hline
\end{tabular}\vspace{-1em}
\end{table*}

\subsection{Retrieval}
We evaluate \name\ on the GeoSound and SoundingEarth benchmarks, comparing its performance with both GeoCLAP~\cite{khanal2023soundscape} and the previous state-of-the-art soundscape mapping method, PSM~\cite{khanal2024psm}. Following the standard evaluation protocol for these benchmarks, we assess \name\ on the image--audio retrieval task using Recall at 10\% (R10\%) and Median Rank (MR). Specifically, R10\% measures the percentage of queries where the ground-truth target is ranked within the top 10\% of the entire test set. Furthermore, to demonstrate \name’s ability to retrieve accurate synthetic captions, we also report results on the image–text retrieval task, where the text corresponds to semantically rich, synthetic captions. 

\subsubsection{Image-Audio Retrieval}
Table~\ref{tab:main} summarizes results for models trained with and without metadata on both the benchmarks with imagery at scale~1. When trained on GeoSound with Bing imagery, \name\ improves I2A-R10\% from 0.423 (PSM) to 0.534 without metadata, and from 0.828 to 0.871 with metadata. Using Sentinel imagery, performance increases from 0.474 to 0.549 without metadata, and from 0.802 to 0.868 with metadata. On SoundingEarth, I2A-R10\% improves from 0.514 to 0.570 without metadata and from 0.563 to 0.626 with metadata. Similar gains are observed for audio-to-image retrieval across both datasets. As shown in supplemental Tables~\ref{tab:supp_SoundingEarth},~\ref{tab:multiscalebingmap}, and~\ref{tab:multiscalesentinel}, these improvements hold consistently across all image scales and retrieval settings.

The results of \name\ demonstrate strong performance in multi-scale satellite image-to-sound retrieval, with clear gains when trained and evaluated using metadata. To analyze the contribution of each metadata component, we perform ablations using individual or subset combinations (supplemental Tables~\ref{tab:metadata_ablation} and~\ref{tab:metadata_ablation2}). The audio source emerges as the most impactful factor, consistent with prior findings in PSM~\cite{khanal2024psm}. This effect of metadata is modest on SoundingEarth, where all samples are from a single source (Aporee:Radio~\cite{aporee}), but pronounced on GeoSound, which aggregates data from four distinct platforms—Freesound~\cite{Freesound}, iNaturalist~\cite{iNaturalist}, Aporee:Radio~\cite{aporee}, and Flickr~\cite{thomee2016yfcc100m}. These sources emphasize different sound types, such as nature sounds from iNaturalist and human activity from Flickr. Consequently, modeling the \textit{audio source} metadata enables \name\ to learn embeddings that account for dataset-specific biases. At inference time, users can also condition retrievals on a selected source, producing metadata-aware soundscape maps tailored to expected sound distributions.

\noindent
\textbf{Composed Retrieval:} The existing state-of-the-art method, PSM, reports cross-modal retrieval results between satellite imagery and audio by incorporating the audio-caption embedding into both modalities during inference. In contrast, we introduce a more realistic variant of this composed-retrieval setting, where the caption embedding is added only to the audio query. This design better reflects practical scenarios in which off-the-shelf audio captioning models~\cite{deshmukh2023pengi, Qwen-Audio} could enhance audio representations but are unavailable for images. More broadly, we evaluate \name\ under two composed-retrieval protocols. Following PSM, we first consider the \textit{Composed (query)} setting, where the caption embedding is added to the query modality (either audio or image). We then propose a fairer variant, \textit{Composed (audio-only)}, in which the caption embedding is applied exclusively to the audio modality. Results for both variants—along with retrieval performance across multiple spatial scales—are provided in Tables~\ref{tab:supp_SoundingEarth},~\ref{tab:multiscalebingmap}, and~\ref{tab:multiscalesentinel}. As shown, \name\ achieves SOTA performance in most settings for satellite–image–to–audio cross-modal retrieval.

\begin{figure*}[t]
    \centering
\includegraphics[width=\textwidth]{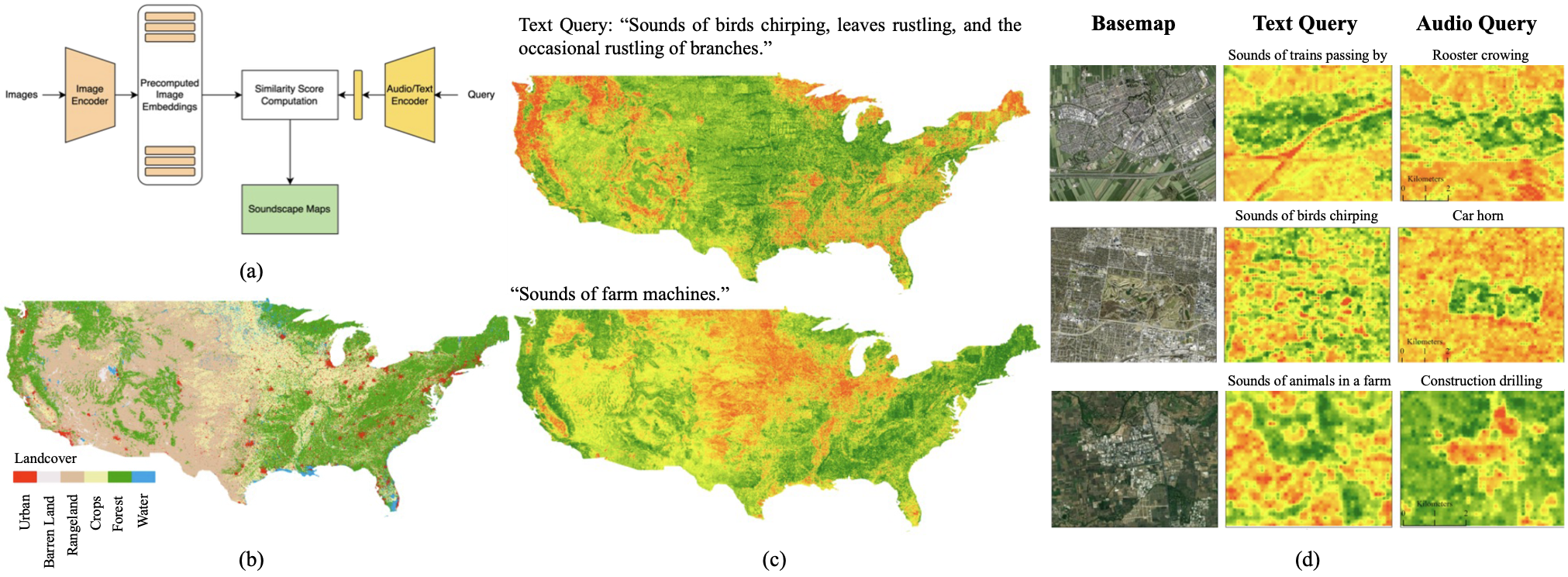}
    \caption{(a) Soundscape mapping framework using \name's encoders. (b) A landcover map for the United States for comparison to soundscape maps. (c) Country-scale soundscape maps created for queries over the USA with a reference land cover map for comparision. (d) City-scale soundscape maps using different queries for cities in the Netherlands (top), the USA (middle), and India (bottom). }
    \label{fig:soundscape-maps}
\end{figure*}

\subsubsection{Image-Text Retrieval}
For evaluation of image-text retrieval performance, the LLaVA-generated soundscape caption serves as our text to be retrieved for an image. Unlike the often noisy or missing text annotations for audios in the dataset, using LLaVA ensures a semantically rich soundscape description for each location.
To compare the performance of \name\ for image-to-text retrieval, we create a strong baseline similar to \name, trained using the same encoders but only on image and image-caption pairs, without any metadata. We refer to this baseline as \texttt{Sat2Text} in our paper. \texttt{Sat2Text} was trained using $\mathcal{L}_{i,t}^{\dagger}$ (Equation \ref{eq:7}) as the training objective. Evaluation of image-text retrieval is conducted using Image-Text Recall@10\% (R10\%) and Median Rank (MR). Additionally, to assess the similarity between the ground-truth image caption and the top-1 retrieved caption, we compute standard machine translation metrics: METEOR, BLEU, and F1 BERTScore (BERT-F1).

Table~\ref{tab:text_retrieval} reports results on the GeoSound dataset with Bing imagery. Averaged across three image scales, \name\ achieves an I2T-R10\% of 0.898 without metadata and 0.923 with metadata, comparable to the \texttt{Sat2Text} baseline (0.914). Caption similarity metrics also show close alignment: METEOR scores of 0.681 for the baseline, 0.664 for \name\ without metadata, and 0.676 with metadata, indicating high semantic consistency (examples in supplemental Figure~\ref{fig:image2text_retrieval_examples}). 

Overall, \name\ accurately retrieves semantically relevant captions, with minimal difference between the metadata and non-metadata settings. This stability is expected since LLaVA-generated captions depend only on image content rather than metadata. The strong retrieval and caption similarity scores further motivate using the top-1 retrieved text as input to text-to-audio generators such as TangoFlux~\cite{hung2024tangoflux}, enabling semantically rich sound synthesis.

\subsection{Soundscape Mapping}
Utilizing the multimodal embedding space of \name\ we can create large-scale soundscape maps for any region. As illustrated in Figure \ref{fig:soundscape-maps} (a), first, the satellite images (Bing or Sentinel) covering the geographic region of interest are downloaded. Then, for the desired scale and metadata settings, image embeddings are computed for each image. Finally, cosine similarity scores between the query embedding and all of the pre-computed image embeddings are used to create soundscape maps.

Using our best-performing model trained with Bing imagery, Figure \ref{fig:soundscape-maps} (c) was created with Bing images at scale 1 ($300 \times 300$ px, 0.6m GSD, zoom level 18) covering the USA, using two textual prompts: \textit{sounds of birds chirping, leaves rustling, and the occasional rustling of branches} and \textit{sounds of farm machines}. The highlighted regions correspond to forest and cropland, respectively, in the reference land-cover map in Figure~\ref{fig:soundscape-maps} (b). Following the same procedure, regional-scale soundscape maps were created using textual and audio prompts from \texttt{Freesound}, as shown in Figure \ref{fig:soundscape-maps} (d). The textual query \textit{sounds of trains passing by on the tracks} highlights the railway track but not the adjacent neighborhood; an audio clip of a car horn activates the urban area but not the large park within it; and \textit{construction drilling} activates urban areas while \textit{sounds of animals on a farm} activates non-urban areas. These results demonstrate \name's ability to generate semantically meaningful soundscape maps.

\subsection{Codebook-guided Local Alignment}
\label{sec:codebook_alignment}

\begin{figure}[t]
    \centering
\includegraphics[width=.9\columnwidth]{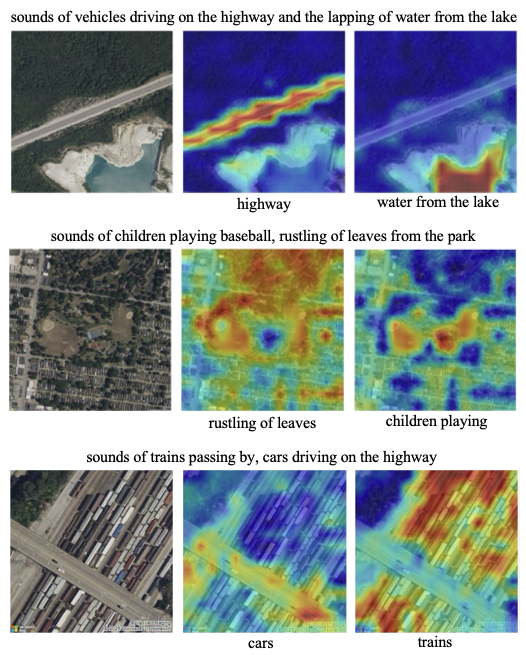}
    \caption{Alignment between patches in a single image and soundscape concepts in textual query.}
    \label{fig:fine-grained-alignment}
\end{figure}

\name\ learns a shared codebook of soundscape concepts that serves as a discrete interface between modalities. Each codebook entry acts as an anchor connecting localized visual patterns and recurring acoustic or textual semantics.
Although the codebook is not trained with explicit concept supervision, these associations emerge naturally through the contrastive objective: during training, image patches, audio segments, and text tokens that frequently co-occur are encouraged to map to the same subset of codebook entries.
As a result, the codebook discretizes the multimodal embedding space into semantically coherent soundscape concepts—such as ``traffic,'' ``water,'' or ``birdsong''—without requiring labeled concepts or patch-level annotations.

Once trained, this discrete structure enables fine-grained local alignment between visual regions and textual or acoustic cues. Given token-level embeddings for a soundscape query ($h^{t} \in \mathbb{R}^{N^t \times d}$), we compute the inner product between $h^{t}$ and the learned codebook ($C \in \mathbb{R}^{M \times d}$), obtaining attention scores between words and codebook entries. For a target word, we select the concept with the highest attention score and use its index to retrieve attention scores for all image patches associated with that concept. For multi-word phrases, scores for each constituent word are averaged across patches to yield the grounded attention map.

Compared to prior frameworks \cite{khanal2023soundscape, khanal2024psm}, which learn only global image-level embeddings, this design provides a major interpretability advantage. Global embeddings mix multiple sound sources within a single vector, making it impossible to visualize where in an image a particular acoustic cue is represented. In contrast, \name’s codebook acts as a shared semantic vocabulary: because individual entries correspond to specific latent soundscape concepts, the model can localize these concepts at the patch level, producing hyper-localized, concept-specific maps that reveal which visual regions drive similarity to a given sound or textual phrase (Figure ~\ref{fig:fine-grained-alignment}). This capability transforms soundscape mapping from a global retrieval problem into a spatially interpretable one—linking fine-grained environmental elements to their corresponding acoustic signatures.

\section{Location-based Soundscape Synthesis}
Recent advances in text-to-audio generation have made it possible to imagine producing the sounds of any location directly from imagery. However, such generative models require substantial computational resources and are difficult to deploy at scale. In \name, we take a different approach: instead of generating new audio at inference time, we leverage richly descriptive synthetic captions and a powerful multimodal retrieval model to \emph{find} the most representative sound for any given location. By training on both human-annotated and LLaVA-generated captions, we construct a large, diverse corpus of soundscapes that can be efficiently queried through our shared embedding space. This enables realistic, location-conditioned audio synthesis without performing generation at inference time. To contextualize our approach, we first describe a cascaded generative baseline and then present our retrieval-based alternative.

\textbf{Cascaded Generative Framework.} This baseline combines image-to-text and text-to-audio generation. Given a satellite image, we query a vision–language model (LLaVA) to produce a detailed description of expected sounds, then turn that caption into sound using the state-of-the-art text-to-audio generator TangoFlux~\cite{hung2024tangoflux}. While this requires no task-specific training, it has substantial computational cost (over 100 seconds of CPU time for a single location) making it unsuitable for interactive applications.

\textbf{Retrieval-based Framework.} In contrast, \name's embedding space and gallery of rich synthetic audio captions support an efficient retrieval-based alternative. Rather than generating captions and audio on demand, we precompute a global gallery of synthetic audio (generated once using TangoFlux) for each of our synthetically generated text captions. At inference, \name\ then retrieves the most semantically aligned caption–audio pair from this gallery (Figure~\ref{fig:image2text_retrieval_examples}), effectively ``playing back’’ the most representative sound for the queried location. This process requires only lightweight embedding lookup (0.14 TFLOPS, $<$1s latency on CPU) while achieving human ratings comparable to the fully generative baseline. We demonstrate this framework through a demo available in our code repository\footnote{\url{https://github.com/mvrl/sat2sound}}.

{\setlength{\abovedisplayskip}{0pt}
\setlength{\belowdisplayskip}{0pt}
\begin{table}[t]
\centering
\small
\setlength{\tabcolsep}{3pt} %
\caption{Comparison of soundscape synthesis methods.}
\label{tab:human_study}
\begin{tabular}{l|c|c|c}
\hline
Approach & Score & \#Params & TFLOPS \\
\hline
Generative & 3.77$\pm$0.51 & 7.57B & 49.03 \\
Retrieval & 3.52$\pm$0.48 & 130M & 0.14 \\
\hline
\end{tabular}
\end{table}
}

To evaluate both frameworks, we conducted a perceptual study in which 16 participants rated the plausibility of generated sounds for 20 geographically diverse locations. Participants scored how likely each audio sample was to be heard at the shown location on a scale from 1–5. As summarized in Table~\ref{tab:human_study}, our retrieval-based framework achieved similar perceptual ratings to the cascaded generative approach, despite requiring two orders of magnitude less compute (130M vs. 7.6B parameters; 0.5s vs. 102s CPU latency). Retrieval-based synthesis thus offers comparable perceptual quality at a fraction of the cost, enabling applications impractical for GPU-dependent generative models — such as interactive web exploration, mobile augmented reality, and large-scale educational tools for acoustic ecology and urban planning.

\section{Conclusion}

We introduced \name, a unified multimodal framework for learning geospatial sound representations by jointly aligning audio, textual descriptions, and satellite imagery through a shared codebook of soundscape concepts. By incorporating both human-annotated and VLM-generated captions, \name\ learns rich correspondences between visual and acoustic patterns, enabling interpretable, fine-grained soundscape mapping at global scale.

Beyond mapping, we demonstrated that this unified embedding space supports efficient, retrieval-based sound synthesis, offering realistic, location-conditioned audio without expensive generative inference. Together, these capabilities establish \name\ as a practical foundation for large-scale environmental monitoring, interactive auditory exploration, and future research on multimodal understanding of place.

\section{Acknowledgements}
This research used the TGI RAILs advanced compute and data resource, which is supported by the National Science Foundation (award OAC-2232860) and the Taylor Geospatial.
{
\small\bibliographystyle{ieeenat_fullname}\bibliography{main}
}

\clearpage
\setcounter{page}{1}
\maketitlesupplementary

\section{Experimental Details}
\label{sec:expr_details}
\textbf{Datasets:} We experiment with two datasets: \textit{GeoSound} and \textit{SoundingEarth}. \textit{GeoSound} contains $294019/5000/9931$ train/validation/test samples and uses both $0.6m$ GSD (Ground Sample Distance) \textit{Bing} image tiles ($1500 \times 1500$) and $10m$ GSD \textit{Sentinel-2} image tiles ($1280 \times 1280$). \textit{SoundingEarth} with $0.2m$ GSD \textit{Google Earth} satellite image tiles of size ($1024 \times 1024$) contains $41469/3242/5801$ train/validation/test samples.\\
\noindent
\textbf{Input Processing:} We process our three input modalities: audio, text, and image as follows:\\
\textit{Audio:}We convert all input audio to mono, randomly sample a 10-second segment, and resample it to 32,000 Hz. The audio then undergoes STFT (window size 1024, hop length 320), followed by conversion to a 64-band Mel spectrogram (50–14,000 Hz), yielding a tensor of shape $1001 \times 64$, where $N^a = 1001$ denotes the number of temporal frames and $F = 64$ denotes the number of Mel frequency bins.\\
\textit{Text:} Both audio captions and image captions are tokenized using the \texttt{google/flan-t5-large} tokenizer with \texttt{model\_max\_length} of $512$.\\
\textit{Image:} For the GeoSound dataset, we center-crop satellite images using a scale factor ($s$) from \(\{1, 3, 5\}\), multiplied by the source-specific tile sizes (256\,px for Sentinel-2 and 300\,px for Bing). During training, to learn a unified multi-scale embedding space, we uniformly sample $s$ from \(\{1, 3, 5\}\). For the SoundingEarth dataset, we apply a single-scale center crop of 256\,px (i.e., scale \(=1\)). In both cases, the cropped images are resized to \(224\times224\) pixels and augmented with color jitter and normalization during training. The image encoder patchifies the image with a \(16 \times 16\) patch size, producing 196 tokens (\(N^{i,s}\)).\\
\noindent
\textbf{Metadata:} Following PSM \cite{khanal2024psm}, \name\ is also trained with metadata (geolocation, month, hour, audio source, and audio caption source) in addition to satellite imagery and associated audio and text. For the GeoSound dataset used in our work, geotagged audio was collected from four sources: Freesound, Aporee, iNaturalist, and Flickr. Each metadata component is embedded into a 1024-dimensional vector and fused using \name's transformer-based metadata fusion module. To prevent overfitting, we apply a dropout rate of 0.5, independently dropping each metadata component during training.\\
\textbf{Audio Captions:} The audio caption can either come from the user-uploaded textual description or be generated using recent SOTA audio-to-text generation models such as Pengi \cite{deshmukh2023pengi} or Qwen-Audio \cite{Qwen-Audio}, with the caption selection based on the caption's CLAP score \cite{wu2023large} with the ground-truth audio. For the \textit{GeoSound} dataset, this resulted in 58.7\% of audio captions from Pengi, 23.8\% from Qwen-Audio, and 17.5\% from human-annotated text.\\
\noindent
\textbf{Image Captions:} For the cropped satellite images at each scale, we generate detailed soundscape captions using LLaVA \cite{liu2024visual}, a powerful open-source Vision Language Model that has been proven effective in captioning satellite images.  Specifically, we query \texttt{llava-hf/llava-1.5-7b-hf} on \texttt{HuggingFace} using the following prompt:
\textit{``What types of sounds can we expect to hear from the location captured by this aerial view image? Describe in up to two sentences."}\\
\noindent
\textbf{Encoders:} Following \cite{khanal2024psm}, we fine-tune the pre-trained checkpoint for the SATMAE-Base \cite{satmae2022} to encode satellite imagery while updating its positional embeddings with scale-aware GSDPE \cite{reed2023scale} to encode the scale of the satellite image. For audio, we fine-tune the pre-trained audio encoder of MGA-CLAP \cite{li2024advancing}, which generates frame-level audio embeddings. The textual modality is processed using a frozen FLAN-T5 \cite{roberts2023scaling} model, which extracts token embeddings from texts for each sample.\\
\noindent
\textbf{Hyper-parameters:} We set the embedding dimension of \name~($d$) to 1024 and the number of concepts in the codebook ($M$) to 16000. We train our model using the \texttt{AdamW} optimizer with cosine-annealing with warm restarts as the learning rate scheduler with the following parameters: learning rate of 5e-5, weight decay of 0.2, and betas of (0.9, 0.98). We set the pseudo-positives contribution to loss ($\alpha$) in Equation \ref{eq:7} to 0.1. We train \name\ for 20 epochs with train-batch size ($B$) of 128. For evaluation, we select the checkpoint that achieves the best I2A-R@10\% performance on our validation set.\\
\noindent
\textbf{Compute Infrastructure:} All experiments were conducted on an \texttt{NVIDIA H100 80GB} GPU, using 16 workers to enable faster data loading. We employed full-precision training throughout. \\
\noindent
\textbf{Human Study:} In this study, 16 participants were shown a Bing satellite image at scale 1 for 20 locations on Earth. These 20 locations were selected by clustering SatCLIP's \cite{klemmer2023satclip} geolocation embeddings of all the samples in our gallery, with the centroid of each cluster serving as a test location. Each satellite image was paired with two $10$-second synthetic audios generated using TangoFlux\cite{hung2024tangoflux} with the inference parameters: \texttt{steps = 50} and \texttt{guidance = 4.5}. One audio was generated using the top-1 retrieved image caption by \name, and the second using the directly generated LLaVA caption passed to TangoFlux.

\section{Ablation Studies}
\label{sec:ablation}
\begin{table*}[h]
\centering
\small
\setlength{\tabcolsep}{6pt}
\caption{Ablation of different loss components. Evaluated for cross-modal image-to-audio retrieval on GeoSound with Bing imagery at scale 1.}
\label{tab:loss_ablation}
\begin{tabular}{c|c|c|c|c|c|c}
\hline
trimodal & L(a+c) & L(i,t) & I2A-R@10\% & I2A-MdR & A2I-R@10\% & A2I-MdR \\
\hline
\checkmark &  & & 0.866  & 192 & 0.871  & 179  \\

\checkmark &   & \checkmark & 0.852 & 206 & 0.852 & 203  \\

\checkmark &  \checkmark &  & 0.873 & 182 & 0.876 & 169  \\

\checkmark &  \checkmark & \checkmark & 0.871 & 168 & 0.875 & 164 \\
\hline
\end{tabular}
\end{table*}

\begin{table*}[h]
\centering
\small
\setlength{\tabcolsep}{6pt}
\caption{Ablation of different loss components. Evaluated for cross-modal composite audio-to-image retrieval on GeoSound with Bing imagery at scale 1.}
\label{tab:loss_ablation_composed}
\begin{tabular}{c|c|c|c|c|c|c}
\hline
trimodal & L(a+c) & L(i,t) & I2A-R@10\% & I2A-MdR & A2I-R@10\% & A2I-MdR \\
\hline
\checkmark &  & & 0.935  & 88 & 0.949  & 75  \\

\checkmark &   & \checkmark & 0.922 & 105 & 0.937 & 94  \\

\checkmark &  \checkmark &  & 0.960 & 71 & 0.962 & 62  \\

\checkmark &  \checkmark & \checkmark & 0.955 & 70 & 0.958 & 64 \\
\hline
\end{tabular}
\end{table*}

\begin{table*}[h]
\centering
\small  %
\setlength{\tabcolsep}{4pt}  %
\caption{Metadata ablation to evaluate \name~models trained on GeoSound dataset with satellite imagery at scale $1$.}
\label{tab:metadata_ablation}
\begin{tabular}{l|c|c|c|c|c|cccc}
\hline
Imagery & latlong & month & time & a-source & c-source & I2A-R10\% & I2A-MR & A2I-R10\% & A2I-MR \\
\hline
Sentinel & $\checkmark$ &  &   &  &  & 0.603 & 613 & 0.627 & 566 \\
Sentinel &  & $\checkmark$ &  &  &  & 0.585 & 682 & 0.604 & 613 \\
Sentinel &  &  & $\checkmark$ &  &  & 0.641 & 537 & 0.669 & 477 \\
Sentinel &  &  &  & $\checkmark$ &  & \textbf{0.805} & \textbf{318} & \textbf{0.808} & \textbf{306} \\
Sentinel &  &  &  &  & $\checkmark$ & 0.562 & 763 & 0.590 & 672 \\
\hline
Bing & $\checkmark$ &  &  &  &  & 0.627 & 547 & 0.644 & 499 \\
Bing &  & $\checkmark$ &  &  &  & 0.577 & 697 & 0.594 & 638 \\
Bing &  &  & $\checkmark$ &  &  & 0.629 & 564 & 0.651 & 521 \\
Bing &  &  &  & $\checkmark$ &  & \textbf{0.787} & \textbf{325} & \textbf{0.793} & \textbf{320} \\
Bing &  &  &  &  & $\checkmark$ & 0.551 & 785 & 0.566 & 742 \\
\hline
\end{tabular}
\end{table*}

\begin{table*}[h]
\centering
\small
\setlength{\tabcolsep}{6pt}
\caption{Composite metadata ablation to evaluate \name~models trained on GeoSound dataset with satellite imagery at scale $1$.}
\label{tab:metadata_ablation2}
\begin{tabular}{c|ccccc|cccc}
\hline
Imagery & a-source & time & latlong & month & c-source & I2A-R10\% & I2A-MR & A2I-R10\% & A2I-MR \\
\hline
Sentinel & \checkmark & & & & & 0.805 & 318 & 0.808 & 306 \\
Sentinel & \checkmark & \checkmark & & & & 0.819 & 295 & 0.820 & 278 \\
Sentinel & \checkmark & \checkmark & \checkmark &  & & 0.850 & 239 & 0.851 & 227 \\
Sentinel & \checkmark & \checkmark & \checkmark & \checkmark &  & 0.862 & 200 & 0.865 & 192 \\
Sentinel & \checkmark & \checkmark & \checkmark & \checkmark & \checkmark & \textbf{0.868} & \textbf{191} & \textbf{0.872} & \textbf{183} \\
\hline
Bing & \checkmark & & & & & 0.787 & 325 & 0.793 & 320 \\
Bing & \checkmark & \checkmark & & & & 0.807 & 289 & 0.811 & 283 \\
Bing & \checkmark & \checkmark & \checkmark &  & & 0.854 & 209 & 0.857 & 202 \\
Bing & \checkmark & \checkmark & \checkmark & \checkmark &  & 0.866 & 175 & 0.871 & 169 \\
Bing & \checkmark & \checkmark & \checkmark & \checkmark & \checkmark & \textbf{0.871} & \textbf{168} & \textbf{0.875} & \textbf{164} \\
\hline
\end{tabular}
\end{table*}

\subsection{Loss Ablation}
\label{sec:loss_ablation}
\begin{figure*}[h]
    \centering
\includegraphics[scale=0.09]{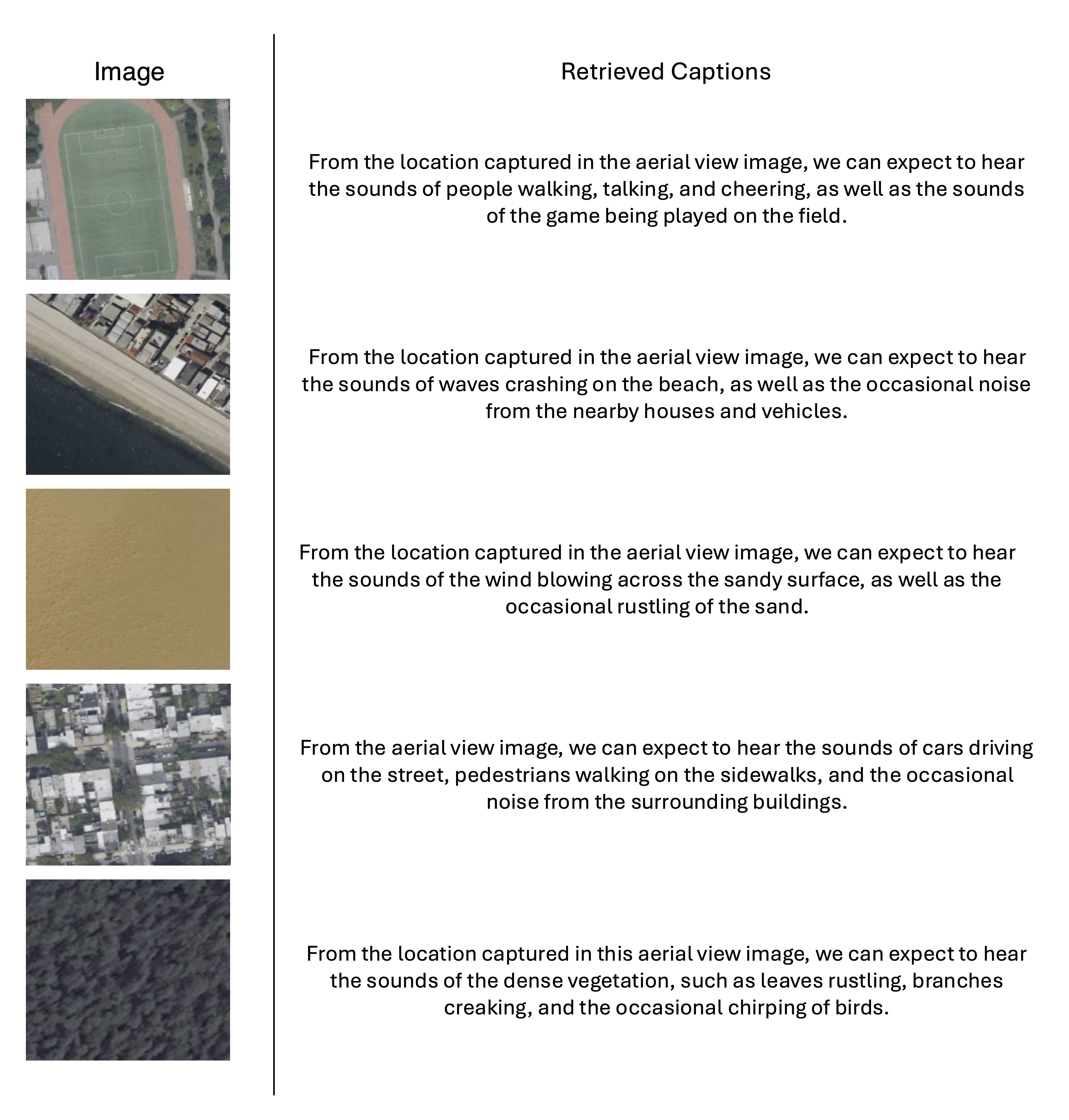}
    \caption{Examples of Top-1 retrieved LLaVA captions for a Bing image by \name\ from our gallery, which is the \textit{test}-set of the GeoSound dataset.}
    \label{fig:image2text_retrieval_examples}
\end{figure*}

We conduct an ablation study on different components of the loss to assess their impact on the overall training objective (Equation \ref{eq:9}). We observe that the addition of the composite audio-based loss ($\mathcal{L}_{i,a+c}^{\dagger}$) slightly improves the performance of the standard audio-image cross-modal retrieval as observed in Table \ref{tab:loss_ablation} and noticeably improves for composed audio-image cross modal retrieval as observed in Table \ref{tab:loss_ablation_composed}. Furthermore, the inclusion of an additional image-text loss ($\mathcal{L}_{i,t}^{\dagger}$) does not degrade performance and provides the benefit of accurate retrieval of text as reflected in Table \ref{tab:text_retrieval}, Figure \ref{fig:image2text_retrieval_examples}, and our demo available in our code repository\footnote{\url{https://github.com/mvrl/sat2sound}}.

\subsection{Metadata Ablation}
\label{sec:metadata_ablation}

This experiment is designed to quantify the impact of different metadata components on the cross-modal retrieval performance of \name. To this end, we evaluate our best-performing Sat2Sound models—trained with either Bing or Sentinel imagery—under varying combinations of metadata availability at inference time. As observed in Table \ref{tab:metadata_ablation}, the most contributing metadata is \textit{audio source}. This is consistent with results in prior work, PSM \cite{khanal2024psm}. 

In addition to the independent metadata ablation presented in Table~\ref{tab:metadata_ablation}, we conduct a more detailed analysis of \emph{composite} metadata combinations, assuming the availability of the \textit{audio source} metadata as a base component (Table~\ref{tab:metadata_ablation2}). This experiment evaluates the incremental contribution of time, geolocation, and month metadata when added cumulatively. The results show a clear pattern: model performance improves consistently as more metadata components are incorporated. Notably, for Bing imagery, the I2A-R@10\% increases from 0.787 (only \textit{audio source}) to 0.866 when time (hour and month) and geolocation (latitude, longitude) are added—demonstrating the benefit of simple, readily available metadata. In contrast, the performance gain from incorporating the \textit{audio caption source} is modest, suggesting that any audio captioning model or user-written text can be used, as long as a reasonable audio caption is provided to query our model.

\subsection{Codebook Size Ablation}
\label{sec:codebook_ablation}
\begin{table}[h]
\centering
\small
\setlength{\tabcolsep}{6pt}
\caption{Codebook ablation for Image-Text retrieval on GeoSound dataset with Bing imagery (scale=1) and corresponding image captions.}
\label{tab:codebook_ablation_text}
\begin{tabular}{l|cc|cc}
\hline
Codebook Size & I2T-R10\% & I2T-MR & T2I-R10\% & T2I-MR \\
\hline
4000        & 0.905 & 167 & 0.915 & 145 \\
8000        & 0.899 & 163 & 0.917 & 146 \\
16000       & 0.908 & 160 & 0.914 & 136 \\
32000       & 0.902 & 165 & 0.914 & 148 \\
\hline
\end{tabular}
\end{table}

\begin{table}[h]
\centering
\small
\setlength{\tabcolsep}{6pt}
\caption{Codebook ablation for Image-Audio retrieval on GeoSound dataset with Bing imagery (scale=1) and corresponding audio.}
\label{tab:codebook_ablation_audio}
\begin{tabular}{l|cc|cc}
\hline
Codebook Size & I2A-R10\% & I2A-MR & A2I-R10\% & A2I-MR \\
\hline
4000        & 0.868 & 167 & 0.870 & 161 \\
8000        & 0.875 & 171 & 0.876 & 163 \\
16000       & 0.871 & 168 & 0.875 & 164 \\
32000       & 0.874 & 164 & 0.879 & 160 \\
\hline
\end{tabular}
\end{table}

We conduct an ablation of the codebook size of our framework. As seen in Tables \ref{tab:codebook_ablation_text} and \ref{tab:codebook_ablation_audio}, performance remains fairly consistent across different codebook sizes. We speculate that the sparsification operation \cite{martins2016softmax} of the attention weights (Equation \ref{eq:4}) encourages our framework to select only the relevant concepts, making the framework independent of the codebook size.

Our choice of codebook-based learning is motivated by the intuition that a fixed set of soundscape concepts can be shared across modalities. This approach offers a more interpretable way to align local features between the satellite image and corresponding soundscape elements. As shown in Figure~\ref{fig:fine-grained-alignment}, this local alignment serves as a valuable byproduct from an interpretability perspective, further discussed in Section~\ref{sec:Analyzing codebook}.

\section{Simpler Baselines}

\label{sec:simpler_baselines}
\begin{table}[h]
\centering
\small
\setlength{\tabcolsep}{6pt}
\caption{Image-Text retrieval comparison with additional baselines. Results on GeoSound with Bing Imagery (scale=1).}
\label{tab:image_text_baselines}
\begin{tabular}{l|cccc}
\hline
Model & I2T-R10\% & I2T-MR & T2I-R10\% & T2I-MR \\
\hline
CLIP \cite{radford2021learning}            & 0.528 & 1420 & 0.461 & 1999 \\
SigLIP \cite{zhai2023sigmoid}          & 0.340 & 3307 & 0.368 & 2652 \\
SigLIP2 \cite{tschannen2025siglip}         & 0.449 & 2641 & 0.397 & 2400 \\
Ours(w/o meta)  & 0.881 & 183  & 0.900 & 166  \\
Ours(w meta)    & \textbf{0.908} & \textbf{160}  & \textbf{0.914} & \textbf{136}  \\
\hline
\end{tabular}
\end{table}

\begin{table}[h]
\centering
\small
\setlength{\tabcolsep}{6pt}
\caption{Image-Audio retrieval comparison with additional baselines. Results on GeoSound with Sentinel Imagery (scale=1).}
\label{tab:image_audio_baselines}
\begin{tabular}{l|cccc}
\hline
Model & I2A-R10\% & I2A-MR & A2I-R10\% & A2I-MR \\
\hline
ImageBind \cite{girdhar2023imagebind}       & 0.214 & 3675 & 0.231 & 3541 \\
TaxaBind \cite{sastry2025taxabind}        & 0.235 & 3448 & 0.250 & 3400 \\
Ours(w/o meta)  & 0.549 & 802  & 0.556 & 778  \\
Ours(w meta)    & \textbf{0.868} & \textbf{191}  & \textbf{0.872} & \textbf{183}  \\
\hline
\end{tabular}
\end{table}

In this section, we compare the performance of \name~ with existing off-the-shelf multimodal embedding spaces. As shown in the image-text cross-modal retrieval results (Table \ref{tab:image_text_baselines}), existing pre-trained image-text models underperform compared to \name. We attribute this to the mismatch between the soundscape descriptions generated by LLaVA from satellite images and the textual data these models were originally trained on. In contrast, Sat2Sound is explicitly trained on these captions, giving it a clear advantage and resulting in significantly better performance than the compared vision-language baselines. A similar trend is observed in Table \ref{tab:image_audio_baselines} for image-audio cross-modal retrieval. These findings underscore the limitations of existing state-of-the-art multimodal embedding spaces for soundscape mapping, highlighting the need for a specialized framework like \name~ tailored to this task.

\section{Multi-scale Cross-Modal Retrieval}
\label{sec:multiscale_retrieval}
\name~ is trained on multi-scale satellite imagery for the GeoSound dataset. The results presented in the main paper are for satellite imagery at scale $1$. In this section, we present results for two additional scales: $3$ and $5$, using both Sentinel and Bing imagery from the GeoSound dataset. Additionally, for both datasets (GeoSound and SoundingEarth), we provide results for composed retrieval settings where the audio caption embedding is added either only to the audio query embedding (indicated as \textit{audio} in the tables) or to both the audio query and image query embeddings (indicated as \textit{query} in the tables), as done in PSM \cite{khanal2024psm}. As observed in Tables \ref{tab:supp_SoundingEarth}, \ref{tab:multiscalebingmap}, and \ref{tab:multiscalesentinel}, \name\ outperforms existing baselines by a noticeable margin in almost all of the settings. 
\begin{table*}[h]
\centering
\small
\setlength{\tabcolsep}{6pt}
\caption{Image-Audio retrieval results for SoundingEarth with different composed audio-image settings.}
\label{tab:supp_SoundingEarth}
\begin{tabular}{l|c|c|c|c|c}
\hline
Method & Composed & I2A-R10\% & I2A-MR & A2I-R10\% & A2I-MR \\
\hline
\multicolumn{6}{c}{\textit{Without Metadata}} \\
\hline
GeoCLAP & query & 0.523 & 533 & 0.470 & 641 \\
PSM & query & 0.687 & 234 & 0.560 & 451 \\
Ours & query & \textbf{0.847} & \textbf{94} & \textbf{0.564} & \textbf{448} \\
\hline
GeoCLAP & audio & 0.478 & 624 & 0.470 & 641 \\
PSM & audio & 0.558 & 462 & 0.560 & 451 \\
Ours & audio & \textbf{0.567} & \textbf{443} & \textbf{0.564} & \textbf{448} \\
\hline
\multicolumn{6}{c}{\textit{With Metadata}} \\
\hline
PSM & query & 0.690 & 264 & 0.608 & 371 \\
Ours & query & \textbf{0.855} & \textbf{91} & \textbf{0.862} & \textbf{129} \\
\hline
PSM & audio & 0.606 & 380 & 0.608 & 371 \\
Ours & audio & \textbf{0.855} & \textbf{127} & \textbf{0.862} & \textbf{129} \\
\hline
\end{tabular}
\end{table*}

\begin{table*}[h]
\centering
\small
\setlength{\tabcolsep}{6pt}
\caption{Image-Audio retrieval results for GeoSound with Bing imagery at different scales.}
\label{tab:multiscalebingmap}
\begin{tabular}{l|c|c|c|c|c|c}
\hline
Scale & Method & Composed & I2A-R10\% & I2A-MR & A2I-R10\% & A2I-MR \\
\hline
\multicolumn{7}{c}{\textit{Without Metadata}} \\
\hline
\multirow{9}{*}{1} & GeoCLAP & query & 0.577 & 712 & 0.468 & 1141 \\
& PSM & query & 0.754 & 204 & 0.510 & 952 \\
& Ours & query & \textbf{0.903} & \textbf{82} & \textbf{0.540} & \textbf{836} \\
\cline{2-7}
& GeoCLAP & audio & 0.464 & 1159 & 0.468 & 1141 \\
& PSM & audio & 0.503 & 980 & 0.510 & 952 \\
& Ours & audio & \textbf{0.535} & \textbf{864} & \textbf{0.540} & \textbf{836} \\
\hline
\multirow{9}{*}{3} & GeoCLAP & none & 0.408 & 1441 & 0.420 & 1389 \\
& PSM & none & 0.440 & 1302 & 0.443 & 1266 \\
& Ours & none & \textbf{0.560} & \textbf{777} & \textbf{0.561} & \textbf{779} \\
\cline{2-7} & GeoCLAP & query & 0.577 & 707 & 0.483 & 1056 \\
& PSM & query & 0.753 & 207 & 0.529 & 880 \\
& Ours & query & \textbf{0.908} & \textbf{79} & \textbf{0.567} & \textbf{737} \\
\cline{2-7}
& GeoCLAP & audio & 0.477 & 1092 & 0.483 & 1056 \\
& PSM & audio & 0.523 & 891 & 0.529 & 880 \\
& Ours & audio & \textbf{0.564} & \textbf{751} & \textbf{0.567} & \textbf{737} \\
\hline
\multirow{9}{*}{5} & GeoCLAP & none & 0.409 & 1428 & 0.421 & 1373 \\
& PSM & none & 0.440 & 1302 & 0.448 & 1279 \\
& Ours & none & \textbf{0.564} & \textbf{760} & \textbf{0.559} & \textbf{770} \\
\cline{2-7}
& GeoCLAP & query & 0.581 & 698 & 0.489 & 1036 \\
& PSM & query & 0.753 & 209 & 0.532 & 863 \\
& Ours & query & \textbf{0.910} & \textbf{78} & \textbf{0.567} & \textbf{748} \\
\cline{2-7}
& GeoCLAP & audio & 0.482 & 1071 & 0.489 & 1036 \\
& PSM & audio & 0.528 & 881 & 0.532 & 863 \\
& Ours & audio & \textbf{0.554} & \textbf{764} & \textbf{0.567} & \textbf{748} \\
\hline
\multicolumn{7}{c}{\textit{With Metadata}} \\
\hline
\multirow{6}{*}{1} & PSM & query & 0.901 & 113 & 0.943 & 100 \\
& Ours & query & \textbf{0.970} & \textbf{33} & \textbf{0.958} & \textbf{64} \\
\cline{2-7}
& PSM & audio & 0.935 & 115 & 0.943 & 100 \\
& Ours & audio & \textbf{0.955} & \textbf{70} & \textbf{0.958} & \textbf{64} \\
\hline
\multirow{6}{*}{3} & PSM & none & 0.827 & 266 & 0.832 & 250 \\
& Ours & none & \textbf{0.874} & \textbf{163} & \textbf{0.879} & \textbf{159} \\
\cline{2-7} & PSM & query & 0.900 & 114 & 0.945 & 102 \\
& Ours & query & \textbf{0.972} & \textbf{32} & \textbf{0.960} & \textbf{62} \\
\cline{2-7}
& PSM & audio & 0.936 & 118 & 0.945 & 102 \\
& Ours & audio & \textbf{0.957} & \textbf{66} & \textbf{0.960} & \textbf{62} \\
\hline
\multirow{6}{*}{5} & PSM & none & 0.821 & 281 & 0.826 & 261 \\
& Ours & none & \textbf{0.877} & \textbf{167} & \textbf{0.882} & \textbf{167} \\
\cline{2-7} & PSM & query & 0.896 & 115 & 0.941 & 107 \\
& Ours & query & \textbf{0.972} & \textbf{32} & \textbf{0.963} & \textbf{64} \\
\cline{2-7}
& PSM & audio & 0.929 & 124 & 0.941 & 107 \\
& Ours & audio & \textbf{0.959} & \textbf{68} & \textbf{0.963} & \textbf{64} \\
\hline
\end{tabular}
\end{table*}

\begin{table*}[h]
\centering
\small
\setlength{\tabcolsep}{6pt}
\caption{Image-Audio retrieval results for GeoSound with Sentinel imagery at different scales.}
\label{tab:multiscalesentinel}
\begin{tabular}{l|c|c|c|c|c|c}
\hline
Scale & Method & Composed & I2A-R10\% & I2A-MR & A2I-R10\% & A2I-MR \\
\hline
\multicolumn{7}{c}{\textit{Without Metadata}} \\
\hline
\multirow{9}{*}{1} & GeoCLAP & query & 0.546 & 827 & 0.553 & 804 \\
& PSM & query & 0.803 & 153 & \textbf{0.595} & \textbf{664} \\
& Ours & query & \textbf{0.909} & \textbf{79} & 0.566 & 748 \\
\cline{2-7}
& GeoCLAP & audio & 0.542 & 809 & 0.553 & 804 \\
& PSM & audio & \textbf{0.586} & \textbf{701} & \textbf{0.595} & \textbf{664} \\
& Ours & audio & 0.555 & 765 & 0.566 & 748 \\
\hline
\multirow{9}{*}{3} & GeoCLAP & none & 0.454 & 1200 & 0.456 & 1197 \\
& PSM & none & 0.479 & 1086 & 0.487 & 1042 \\
& Ours & none & \textbf{0.559} & \textbf{776} & \textbf{0.561} & \textbf{763} \\
\cline{2-7} & GeoCLAP & query & 0.542 & 840 & 0.555 & 790 \\
& PSM & query & 0.799 & 159 & \textbf{0.604} & \textbf{657} \\
& Ours & query & \textbf{0.910} & \textbf{81} & 0.577 & 729 \\
\cline{2-7}
& GeoCLAP & audio & 0.548 & 812 & 0.555 & 790 \\
& PSM & audio & \textbf{0.594} & \textbf{676} & \textbf{0.604} & \textbf{657} \\
& Ours & audio & 0.561 & 757 & 0.577 & 729 \\
\hline
\multirow{9}{*}{5} & GeoCLAP & none & 0.458 & 1194 & 0.457 & 1184 \\
& PSM & none & 0.459 & 1172 & 0.465 & 1138 \\
& Ours & none & \textbf{0.545} & \textbf{804} & \textbf{0.560} & \textbf{774} \\
\cline{2-7}
& GeoCLAP & query & 0.542 & 835 & 0.554 & 791 \\
& PSM & query & 0.796 & 158 & \textbf{0.584} & \textbf{711} \\
& Ours & query & \textbf{0.909} & \textbf{82} & 0.566 & 751 \\
\cline{2-7}
& GeoCLAP & audio & 0.550 & 812 & 0.554 & 791 \\
& PSM & audio & \textbf{0.579} & \textbf{720} & \textbf{0.584} & \textbf{711} \\
& Ours & audio & 0.553 & 784 & 0.566 & 751 \\
\hline
\multicolumn{7}{c}{\textit{With Metadata}} \\
\hline
\multirow{6}{*}{1} & PSM & query & 0.872 & 142 & 0.940 & 104 \\
& Ours & query & \textbf{0.972} & \textbf{35} & \textbf{0.959} & \textbf{70} \\
\cline{2-7}
& PSM & audio & 0.931 & 123 & 0.940 & 104 \\
& Ours & audio & \textbf{0.956} & \textbf{78} & \textbf{0.959} & \textbf{70} \\
\hline
\multirow{6}{*}{3} & PSM & none & 0.795 & 306 & 0.800 & 290 \\
& Ours & none & \textbf{0.857} & \textbf{208} & \textbf{0.858} & \textbf{199} \\
\cline{2-7} & PSM & query & 0.870 & 150 & 0.940 & 104 \\
& Ours & query & \textbf{0.970} & \textbf{37} & \textbf{0.955} & \textbf{74} \\
\cline{2-7}
& PSM & audio & 0.929 & 126 & 0.940 & 104 \\
& Ours & audio & \textbf{0.949} & \textbf{83} & \textbf{0.955} & \textbf{74} \\
\hline
\multirow{6}{*}{5} & PSM & none & 0.794 & 316 & 0.794 & 299 \\
& Ours & none & \textbf{0.846} & \textbf{220} & \textbf{0.851} & \textbf{216} \\
\cline{2-7} & PSM & query & 0.868 & 156 & 0.935 & 109 \\
& Ours & query & \textbf{0.969} & \textbf{37} & \textbf{0.954} & \textbf{80} \\
\cline{2-7}
& PSM & audio & 0.926 & 131 & 0.935 & 109 \\
& Ours & audio & \textbf{0.948} & \textbf{88} & \textbf{0.954} & \textbf{80} \\
\hline
\end{tabular}
\end{table*}

\section{Analyzing codebook concepts}
\label{sec:Analyzing codebook}
\begin{figure*}[h]
    \centering
\includegraphics[scale=0.87]{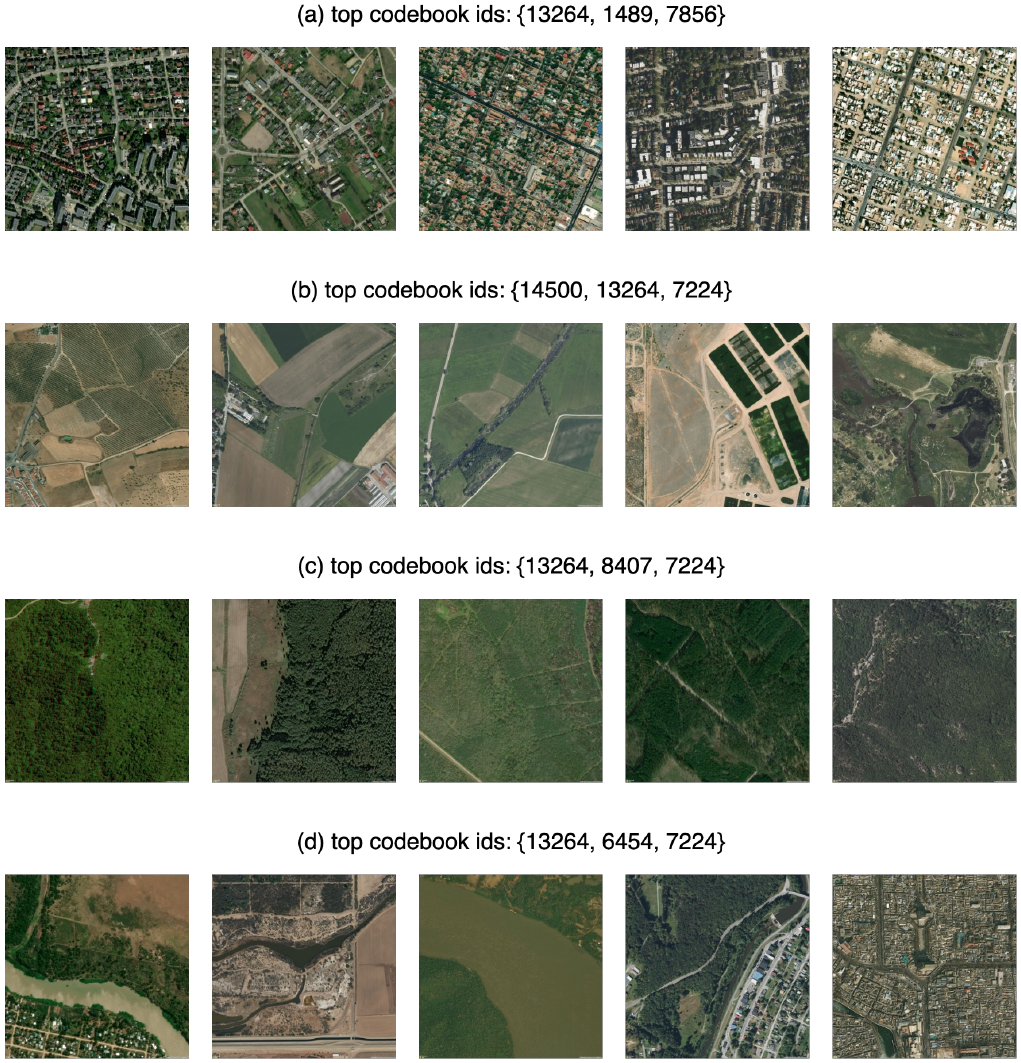}
    \caption{Some example groups from the GeoSound test set are shown, where each group shares a common set of highly activated codebook concepts, reflecting similar soundscapes of specific geographic areas. The samples in (a) correspond to residential soundscapes, (b) reflect the soundscape of open fields, (c) represent forested area soundscapes, and (d) capture the soundscape of landscapes with water bodies.}
    \label{fig:codebook_groups_examples}
\end{figure*}

As illustrated in Figure~\ref{fig:fine-grained-alignment}, the codebook learned by \name~ can be used to generate fine-grained soundscape maps for regions covered by a single satellite image. In this section, we qualitatively explore what the codebook has learned. Specifically, for our gallery of image captions, we first obtain the corresponding codebook attention weights (Equation~\ref{eq:4}) and group together samples that share a similar set of highly activated codebook concepts. For a subset of these groups, we randomly sample examples to examine the behavior and semantic meaning captured by different sets of codebook concepts. Representative samples are visualized in Figure~\ref{fig:codebook_groups_examples}.

\begin{table*}[h]
\centering
\small
\setlength{\tabcolsep}{2pt} %
\renewcommand{\arraystretch}{1.05} %
\begin{minipage}{0.48\textwidth}
\centering
\caption{Linear probing evaluation for audio classification performance across different benchmarks.}
\label{tab:audio_classification}
\begin{tabular}{l|c|c|c}
\hline
Model & BirdCLEF-2022 & BirdCLEF-2023 & BirdCLEF-2024 \\
\hline
CLAP~\cite{laionclap2023} & 42.33 & 32.85 & 39.72 \\
MGA-CLAP~\cite{li2024advancing} & \textbf{56.05} & 44.03 & 47.36 \\
ImageBind~\cite{girdhar2023imagebind} & 47.11 & 37.46 & 45.04 \\
TaxaBind~\cite{sastry2025taxabind} & 52.60 & 42.19 & 49.31 \\
Ours & 54.84 & \textbf{45.69} & \textbf{49.59} \\
\hline
\end{tabular}
\end{minipage}
\hfill
\begin{minipage}{0.48\textwidth}
\centering
\caption{Linear probing evaluation for satellite image classification performance across different datasets.}
\label{tab:sat_classification}
\begin{tabular}{l l|c|c|c}
\hline
Model & ViT & EuroSAT & UC-Merced & RESISC-45 \\
\hline
SatMAE~\cite{satmae2022} & Large & 96.43 & 93.81 & 85.00 \\
SatMAE++~\cite{satmaepp2024rethinking} & Large & 96.57 & 96.43 & 84.13 \\
Ours & Base & 83.83 & 90.95 & 74.40 \\
\hline
\end{tabular}
\end{minipage}
\end{table*}

\section{Linear Probing Experiments}
\name~learns a multimodal embedding space between audio and satellite imagery. We evaluate these embeddings on two downstream tasks: audio classification and satellite image classification, using linear probing on the audio and image embeddings, respectively.

For audio classification, we compare the Sat2Sound audio encoder against four strong baselines across three bird sound classification benchmarks: BirdCLEF-2022, BirdCLEF-2023, and BirdCLEF-2024. The baselines include CLAP~\cite{laionclap2023}, MGA-CLAP~\cite{li2024advancing}, ImageBind~\cite{girdhar2023imagebind}, and TaxaBind~\cite{sastry2025taxabind}. CLAP and MGA-CLAP represent state-of-the-art audio–text models, while ImageBind and TaxaBind align multiple modalities within a shared embedding space. As shown in Table~\ref{tab:audio_classification}, Sat2Sound achieves the best performance on two of the three benchmarks and the second-best on the remaining one. We attribute this to Sat2Sound being trained on the GeoSound dataset, where a significant portion of the samples originate from iNaturalist, providing diverse coverage of bird sounds.
For satellite image classification, we evaluate Sat2Sound image embeddings on EuroSAT, UC-Merced, and RESISC-45, comparing them with SatMAE~\cite{satmae2022} and SatMAE++~\cite{satmaepp2024rethinking}. Sat2Sound performs comparably but does not surpass either baseline, both of which use ViT-Large backbones compared to Sat2Sound’s ViT-Base encoder. Nonetheless, Sat2Sound offers broader multimodal capabilities beyond pure image classification.

\end{document}